\documentclass{article} % For LaTeX2e
\usepackage[preprint]{colm2026_conference}

\usepackage{microtype}
\usepackage{hyperref}
\usepackage{url}
\usepackage{booktabs}

\usepackage{amsmath}
\usepackage{algorithm}
\usepackage{algorithmic}

\usepackage{multirow}

\usepackage{graphicx}

\usepackage{wrapfig}

\usepackage{lineno}

\definecolor{darkblue}{rgb}{0, 0, 0.5}
\hypersetup{colorlinks=true, citecolor=darkblue, linkcolor=darkblue, urlcolor=darkblue}

\title{Diffusion In Diffusion: Reclaiming Global Coherence in Semi-Autoregressive Diffusion}

\author{Linrui Ma, Yufei Cui\thanks{Project Leader}~,~Kai Han\thanks{Corresponding Authors}~\&~Yunhe Wang\footnotemark[2] \\
Noah's Ark Lab, Huawei\\
Montreal, Canada \& Beijing, China \\
\texttt{\{kai.han, yunhe.wang\}@huawei.com} \\
}

\begin{document}

\ifcolmsubmission
\linenumbers
\fi

\maketitle

\begin{abstract}
One of the most compelling features of global discrete diffusion language models is their global bidirectional contextual capability. 
However, existing block-based diffusion studies tend to introduce autoregressive priors, which, while offering benefits, can cause models to lose this global coherence at the macro level.
To regain global contextual understanding while preserving the advantages of the semi-autoregressive paradigm,  we propose \textsc{Diffusion in Diffusion} — a 'draft-then-refine' framework designed to overcome the irreversibility and myopia problems inherent in block diffusion models.
Our approach first employs block diffusion to generate rapid drafts using small blocks, then refines these drafts through global bidirectional diffusion with a larger bidirectional receptive field.
We utilize snapshot confidence remasking to identify the most critical tokens that require modification, and apply mix-scale training to expand the block diffusion model's global capabilities.
Empirical results demonstrate that our approach sets a new benchmark for discrete diffusion models on the OpenWebText dataset.
Using only 26\% of the fine-tuning budget of baseline models, we reduce generative perplexity from 25.7 to 21.9, significantly narrowing the performance gap with autoregressive models.
\end{abstract}

\section{Introduction}
\label{sec:intro}

% Today, language modeling primarily follows two mainstream paradigms: autoregressive (AR) models and discrete diffusion models. AR models \citep{Vaswani+2017, brown2020language} generate tokens sequentially from left to right and achieve efficient inference through key-value (KV) caching during inference. However, they suffer from linearly increasing inference latency. Conversely, discrete diffusion models \citep{austin2021structured, lou2024discrete} enable parallel generation with flexible control over efficiency and quality. Yet, the lack of KV caching hinders efficient inference, and they often underperform AR models on perplexity metrics. Recently, block-diffusion language models (BD3-LMs) \citep{arriola2025block} have emerged as a semi-AR architecture. By executing diffusion in parallel within “blocks” while employing autoregressive sequence generation between blocks, BD3-LMs successfully bridge the gap between the two paradigms: achieving state-of-the-art performance in non-autoregressive methods while effectively leveraging KV caching for inference acceleration.

Today, research on discrete diffusion models \citep{austin2021structured, lou2024discrete} is divided into two major paradigms: global diffusion models and block diffusion models (also known as Semi-Autoregressive diffusion models). Global diffusion models \citep{sahoo2024simple, nie2025large} possess strong global planning capabilities and complete bidirectional context, resulting in text with high global consistency. However, they also prevent the application of Key-Value Cache—the foundation for inference acceleration in modern language models \citep{Vaswani+2017, brown2020language} —leading to an $O(L^2)$ complexity during inference and extremely slow processing. In contrast, block diffusion models \citep{arriola2025block} represent an emerging semi-AR architecture that incorporates autoregressive priors. They enable parallel diffusion execution within “blocks” while employing inter-block autoregression for sequence generation. This enables rapid inference leveraging KV Cache while benefiting from the performance gains of autoregressive priors. However, its macro-level autoregressive paradigm imposes a “shortsightedness,” sacrificing diffusion’s core essence—the global receptive field—and degenerating into a local model. It also introduces irreversibility similar to AR, preventing modifications to already written content.

% However, the standard block diffusion model inevitably introduces myopia and irreversibility due to the inclusion of autoregressive components: the model can only access bidirectional contextual information within a single block, while strictly adhering to an autoregressive causal paradigm between blocks. Furthermore, once diffusion within a block is complete, that block becomes “frozen” and permanently serves as an unupdatable “reference-only” contextual block during subsequent block processing.
% This leads to the same error accumulation problem as AR models. Early generated tokens, even if erroneous or locally inconsistent, become locked and immutable. Consequently, subsequent token generation is affected, preventing the model from correcting these issues.
% Therefore, due to the lack of exposure to the global context during generation, BD3-LM struggles to maintain long-range consistency, resulting in a partial loss of global planning capability.

Current research often seeks a balance between the following trade-offs: stronger global planning capabilities necessitate sacrificing inference speed, while better autoregressive priors and faster inference require sacrificing global perspective. We contend that this dichotomy is not irreconcilable. Future architectures should not be inherently AR models with minor local diffusion biases, but rather novel diffusion frameworks that preserve semi-AR advantages while retaining global vision and planning capabilities. By reintroducing global capabilities into Block Diffusion models, we aim to break the impossible triangle of speed vs. quality vs. global consistency in discrete diffusion, resulting in a unified architecture that combines the strengths of both approaches.

In this study, we propose a novel generative framework called \textsc{Diffusion in Diffusion} (also known as Structured Block Diffusion), designing a multi-stage “draft-then-revise” paradigm to bring back the global ability to the aforementioned semi-autoregressive diffusion models.
Inspired by the human writing process—rapidly drafting a first version followed by holistic revision to ensure coherence—our approach decomposes generation into progressive stages:
During the initial drafting phase, the model employs a small blocksize for fast, efficient generation, focusing first on specific grammar and local content.
In subsequent revision phases, the model re-examines and refines the previously generated sequence using a larger blocksize (global receptive field).
We also innovatively introduce a Remasking mechanism into Block-Diffusion Models, which selectively transforms tokens within the generated sequence back into noise. This enables the model to resample these positions within a more holistic context for optimization.

Our work makes three key contributions:
\begin{enumerate}
\item \textbf{Structural Diffusion Framework}: We propose a multi-stage generative framework that progressively increases the block size through multiple stages. This enables the model to first rapidly generate drafts using semi-regressive methods, then fine-tune content with high quality via bidirectional global diffusion, effectively combining the strengths of both paradigms.
\item \textbf{Snapshot Confidence Remask}: We evaluated multiple inter-stage remask token selection strategies and determined that snapshot confidence remask—based on token generation-time confidence—is the most effective approach, outperforming methods like Post-hoc likelihood evaluation.
\item \textbf{Mix-Scale Training Strategy}: We also propose an enhanced Block Diffusion training approach. By mixing block masks of minimal and maximal scales, the model gains multi-granularity information processing capabilities, enabling both generation and revision functions.
\end{enumerate}
Empirical results demonstrate that our approach sets a new benchmark for discrete diffusion models on the OpenWebText dataset using only \textbf{one-quarter} of the fine-tuning budget of comparable models: generating perplexity drops from 25.7 to \textbf{21.9}, significantly narrowing the gap with autoregressive benchmarks.

\section{Preliminaries}

% \paragraph{Notation}

\subsection{Autoregressive Models}
In a discrete (tokenized) sequence $x = [x_1, \dots, x_L]$ of length $L$, each token $x_i$ is a one-hot vector of model vocabulary $\mathcal{V}$. 
The standard autoregressive model decomposes the joint probability distribution $p(x)$ into a product of conditional probabilities:
\begin{equation}
    \log p_\theta(x) = \sum_{i=1}^{L} \log p_\theta(x_i | x_{<i}),
\end{equation}
Here, $x_{<i}$ denotes the prefix token preceding position $i$. The conditional distribution $p_\theta(x_i | x_{<i})$ is typically modeled via a causal attention mechanism.
Although AR models can generate highly fluent and natural text and support key-value caching, their inherently sequential nature precludes straightforward parallel generation.

\subsection{Block Diffusion Models}
Block Diffusion Language Models (BD3-LMs)\citep{arriola2025block} lie between autoregressive and diffusion paradigms. They predefine a fixed block size $\mathcal{B}$ (assuming $B=L/\mathcal{B}$) and partition the entire sequence into $B$ blocks (denoted as $x^1, \dots, x^B$). Each block is subsequently treated as a ‘metatoken’ at the autoregressive level,
\begin{equation}
    \log p_\theta(x) = \sum_{b=1}^{B} \log p_\theta(x^b | x^{<b}), \label{eq:block_ar}
\end{equation}
Here, $x^b$ denotes the token sequence in the $b$th block of $x$, while $x^{<b}$ represents all preceding blocks. Unlike autoregressive models, the conditional distribution within blocks, $p_\theta(x^b | x^{<b})$, is modeled via a discrete diffusion process.

\paragraph{Forward Masking Process}
Following the masked diffusion framework \citep{austin2021structured, sahoo2024simple}, in the forward process $q(x_t^b | x^b)$, each token in $b$ is independently converted to the special masking token \texttt{[MASK]} (denoted as $m$) via according to transition probabilities derived from a noisy scheduling parameter $\alpha_t \in [0, 1]$:
\begin{equation}
    q(x_t^b | x^b) = \prod_{j=1}^{\mathcal{B}} \left( \alpha_t \mathbb{I}(x_{t,j}^b = x_{j}^b) + (1-\alpha_t) \mathbb{I}(x_{t,j}^b = m) \right),
\end{equation}
where $t$ goes from $0$ to $1$, a larger $t$ denotes a noisier timestep, and $t=1$ indicates full noise, and $x_{j}^b$ represents the $j$-th token within the $b$-th block.

\paragraph{Reverse Denoising Process \& Training}
The objective of the generation process (i.e., the denoising procedure) is to learn to reverse this noise-adding process. The aim is to train the denoising network $x_\theta$ to predict a clean current block $x^b$ based on the context provided by the current noisy block $x_t^b$ (subject to its current level of random noise) and the previously denoised clean blocks $x^{<b}$.
Here, BD3-LM employs a key-value cache to leverage prior blocks $x^{<b}$, whilst simultaneously enabling parallel denoising within blocks and decoding multiple tokens concurrently. Consistent with conventional MDLM practices, the model's training objective is to minimise the negative evidence lower bound (NELBO)\citep{sahoo2024simple}:
\begin{equation}
    \mathcal{L}_{\text{BD}}(x; \theta) = \sum_{b=1}^B \mathbb{E}_{t \sim [0,1]} \mathbb{E}_{q(x_t^b|x^b)} \left[ \frac{\alpha'_t}{1-\alpha_t} \mathcal{L}_{\text{CE}}(p_\theta(\cdot | x_t^b, x^{<b}), x^b) \right]. \label{eq:nelbo}
\end{equation}
This objective effectively trains the model to perform "infilling" or conditional generation for the current block for any noise level given the history.

% \section{Structural Block Diffusion}

% \subsection{Overview of multi-stage generation}

\section{Structural Block Diffusion}
\label{sec:method}

\begin{figure}[t]
\begin{center}
\includegraphics[width=\linewidth]{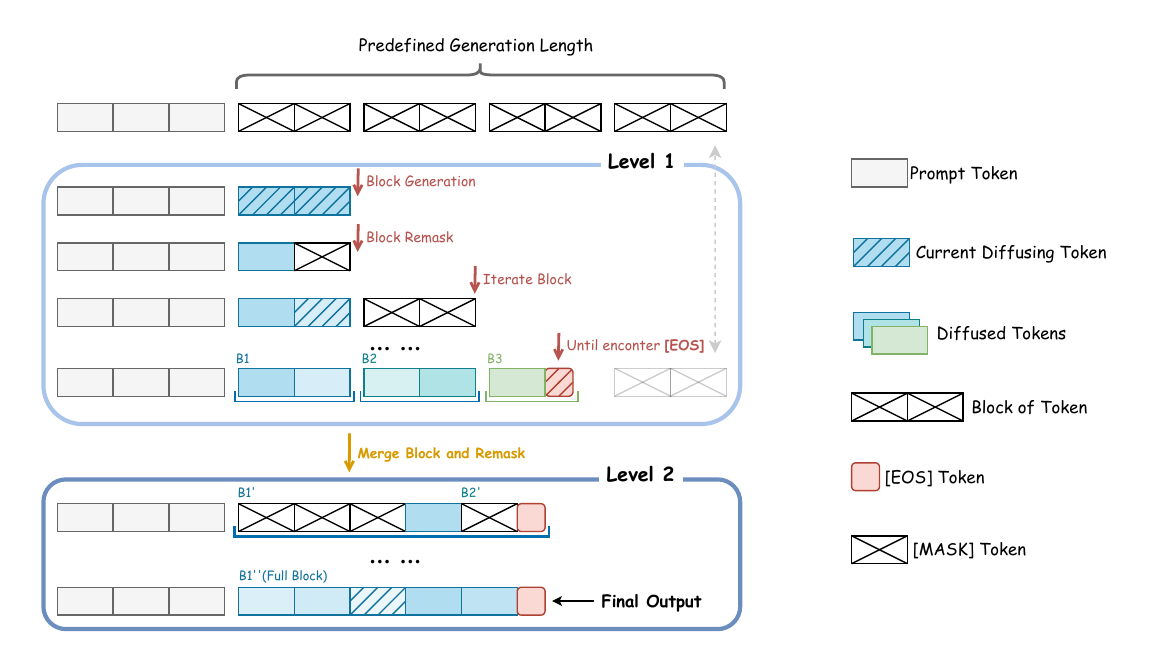}

\end{center}
\caption{Overview of \textsc{Diffusion in Diffusion} method}
\end{figure}

We introduce \textit{Structural Block Diffusion}, a multi-stage generation paradigm designed to overcome the myopic limitation of standard autoregressive block diffusion. 
The core intuition is to decompose the generation process into a drafting phase and one or more revision phases. 
Our approach progressively increases the size of diffusion blocks at each stage, thereby enlarging the bidirectional receptive field for each generation. Between stages, a low-confidence filtering scheme selects the tokens most in need of modification. This partially circumvents the irreversibility inherent in the AR paradigm, enabling the model to revisit earlier generation errors with a more global perspective and greater contextual information.

\subsection{Overview of Multi-Stage Generation}

Formally, we define a generation process consisting of $K$ stages. Let $\mathcal{B}^{(k)}$ denote the block size and $\gamma^{(k)}$ denote the re-masking ratio for stage $k \in \{1, \dots, K\}$. 
We enforce a progressive scaling of the block size such that $\mathcal{B}^{(1)} < \mathcal{B}^{(2)} < \dots < \mathcal{B}^{(K)}=L$. 
In the initial stage ($k=1$), the model generates a "draft" sequence $\hat{x}^{(1)}$ using a small block size (e.g., $\mathcal{B}^{(1)}=4$), which prioritizes local coherence and generation speed but may lack global consistency.

For subsequent refinement stages $k > 1$, the process operates as follows:
\begin{enumerate}
    \item \textbf{Confidence Estimation:} We estimate the confidence score $s_i$ for each token $\hat{x}^{(k-1)}_i$ in the sequence generated from the previous stage. We choose to record the confidence at the moment of every token's sampling and form a confidence trajectory, we name this snapshot confidence(details in Sec. \ref{sec:remasking}).
    \item \textbf{Inter-Stage Remasking:} We identify the set of tokens to retain, $\mathcal{M}_{keep}$, by selecting the top $(1-\gamma^{(k)})$ tokens with the highest confidence scores. The remaining tokens are reset to the mask token $m$.
    % \begin{equation}
    %     x_{init, i}^{(k)} = 
    %     \begin{cases} 
    %     \hat{x}_{i}^{(k-1)} & \text{if } i \in \mathcal{M}_{keep} \\
    %     m & \text{otherwise}
    %     \end{cases}
    % \end{equation}
    \item \textbf{Global Refinement:} We perform block diffusion sampling on $x_{init}^{(k)}$ using the larger block size $\mathcal{B}^{(k)}$. Unlike the empty initialization in Stage 1, Stage $k$ starts from a partially filled sequence. Then we utilize this enhanced global context vision to do infillings for the remasked tokens and thereby improve the overall coherence.
\end{enumerate}

When $\mathcal{B}^{(K)}$ equals the sequence length $L$, the final stage degenerates into a full-sequence masked diffusion process (similar to MDLM \citep{sahoo2024simple}), but with a better partial initialization and a useful structural prior from the first stage's draft. Full procedure is summarized in Algorithm \ref{alg:structural_sampling}.

\begin{algorithm}[t]
\caption{Structural Block Diffusion Sampling}
\label{alg:structural_sampling}
\begin{algorithmic}[1]
\REQUIRE $L$, $x_\theta$, Stages $K$, Block Sizes $\{\mathcal{B}^{(k)}\}_{k=1}^K$, Ratios $\{\gamma^{(k)}\}_{k=2}^K$
\STATE $x \leftarrow m^L$ 
\STATE $\mathcal{S} \leftarrow \mathbf{0}^L$ 
\FOR{$k = 1$ \textbf{to} $K$}
    \STATE $B^{(k)} \leftarrow L / \mathcal{B}^{(k)}$ 
    \IF{$k > 1$}
        \STATE $\mathcal{I} \leftarrow \text{argsort}(\mathcal{S})[1 : \lfloor \gamma^{(k)} L \rfloor]$ 
        \STATE $x_{\mathcal{I}} \leftarrow m$ \COMMENT{Inter-stage Remasking}
    \ENDIF
    \STATE $K_{\text{cache}}, V_{\text{cache}} \leftarrow \emptyset$ 
    \FOR{$b = 1$ \textbf{to} $B^{(k)}$}
        \STATE $x^b \leftarrow x[(b-1)\mathcal{B}^{(k)} : b\mathcal{B}^{(k)}]$
        \STATE $x^b, \mathcal{S}^b \leftarrow \textsc{Sample}(x_\theta, x^b, K_{\text{cache}}, V_{\text{cache}})$ \COMMENT{Block Diffusion}
        \STATE $\_, K^b, V^b \leftarrow x_\theta(x^b)$ 
        \STATE $x \leftarrow x^{1:b-1} \oplus x^b \oplus x^{b+1:B^{(k)}}$ 
        \STATE $\mathcal{S} \leftarrow \mathcal{S}^{1:b-1} \oplus \mathcal{S}^b \oplus \mathcal{S}^{b+1:B^{(k)}}$ 
        \STATE $K_{\text{cache}}, V_{\text{cache}} \leftarrow K_{\text{cache}} \oplus K^b, V_{\text{cache}} \oplus V^b$ 
    \ENDFOR
\ENDFOR
\RETURN $x$
\end{algorithmic}
\end{algorithm}

\subsection{Inter-Stage Remasking Strategy}
\label{sec:remasking}

The transition between stages acts as a quality filter. Instead of carrying over the entire sequence $\hat{x}^{(k-1)}$ to the next stage, we aim to selectively preserve high-quality tokens while resetting uncertain ones to the mask token $m$. This allows the subsequent stage, equipped with a larger receptive field (block size), to focus its generation capacity on "infilling" the problematic regions.
There are two key components in this stage: a confidence-measuring metric and a remasking policy.

\paragraph{Snapshot Confidence Estimation}
To select tokens in need of correction from the draft sequence $\hat{x}^{(k-1)}$ generated in the first stage, an intuitive approach involves performing another forward pass on the generated sequence to obtain likelihood values (posterior confidence). However, this method is highly prone to ‘overconfidence’ – the model repeatedly affirms its own hallucinatory tokens, thereby failing to provide any meaningful guidance.

We regard the uncertainty exhibited by the model at the moment of decision as a free and useful indicator signal, hence we propose Snapshot Confidence. More formally, when the token at position $i$ transitions from $m$ to a specific value $v \in \mathcal{V}$, that token is effectively determined at the particular time step $t$.
Let $p_\theta(x_i = v | x_t^b, x^{<b})$ denote the probability predicted by the denoiser at this transition step (where $b$ is the block containing $i$). We define the snapshot confidence $s_i$ for token $i$ as:
\begin{equation}
    s_i = p_\theta(\hat{x}_i^{(k-1)} | x_{t^*}^b, x^{<b}),
\end{equation}
where $t^*$ denotes the diffusion timestep where token $i$ was unmasked. This metric preserves the dynamic uncertainty inherent in the generation process, thereby providing a more reliable indication of potential errors compared to static post-generation evaluation.

\paragraph{Ratio-based Masking Policy}
After gaining the snapshot confidence trajectory $\mathcal{S} = [s_1, \dots, s_L]$, we use a masking ratio $\gamma^{(k)} \in [0, 1]$to determine in high level how many tokens to be revised. 
This ratio controls the extent of revision: a low $\gamma^{(k)}$ indicates trusting if the stage 1 draft and only revising a minority of tokens, while a high $\gamma^{(k)}$ indicates introducing more global revision forces. Moreover, this $\gamma^{(k)}$ also acts as a user-defined glider controlling the tradeoff between Performance and Efficiency.

We first sort the confidence trajectory $\mathcal{S}$ in an ascent order, then the set of indices to be masked, $\mathcal{I}$ is chosen from the bottom $\gamma$-quantile.
% \begin{equation}
%     \mathcal{I} = \{ i \mid \mathcal{R}(i) \le \lfloor \gamma^{(k)} \cdot L \rfloor \}.
% \end{equation}
Then the input sequence for the subsequent stage $k$, denoted as $x_{init}^{(k)}$, is constructed by remasking the selected tokens at position $\mathcal{I}$ to the mask token $m$:
\begin{equation}
    x_{init, i}^{(k)} = 
    \begin{cases} 
    m & \text{if } i \in \mathcal{I} \\
    \hat{x}_i^{(k-1)} & \text{otherwise}
    \end{cases}
\end{equation}
By remasking the lowest-confidence tokens from the sequence generated in the previous stage, we pre-construct a structured skeleton for the subsequent phase. This enables the effective correction of local inconsistencies through conditional filling operations, leveraging more reliable global dependencies.

\subsection{Mixed-Scale Training Objective}
\label{sec:mixed_training}

Standard training of block diffusion models typically fixes the block size $\mathcal{B}$ as a static hyperparameter during fine-tuning (e.g., $\mathcal{B}=16$ or $\mathcal{B}=32$) \citep{arriola2025block}. 
While effective for single-stage generation, this approach leads to models that are highly specialized to a specific granularity. 
We observe that a model trained exclusively on small blocks fails to generalize to the global context required for the refinement stage, often exhibiting high perplexity or incoherent infilling when presented with a large block size (e.g., $\mathcal{B}=1024$). 
Meanwhile, training only on large blocks is also inefficient in computation and degrades the model to a standard MDLM, hurting the block diffusion local modeling ability required for the initial drafting phase.

To strike a balance between these two approaches, we propose a hybrid-scale training strategy. We reformulate our optimisation problem by treating the block size as a random variable sampled from the distribution $P_{\mathcal{B}}$. The objective function is thereby transformed into the expected value of the block size:
\begin{equation}
    \mathcal{L}_{\text{Mixed}}(\theta) = \mathbb{E}_{\mathcal{B} \sim P_{\mathcal{B}}} \left[ \mathcal{L}_{\text{BD}}(x; \theta, \mathcal{B}) \right],
\end{equation}
where $\mathcal{L}_{\text{BD}}(x; \theta, \mathcal{B})$ is the block diffusion loss (Eq. \ref{eq:nelbo}) computed using block decomposition size $\mathcal{B}$.

\paragraph{Bimodal Block Distribution}
The design of $P_{\mathcal{B}}$ is crucial for balancing the "drafting" and "refining" capabilities. We employ a bimodal distribution parameterized by a mixing coefficient $\lambda \in [0, 1]$:
\begin{equation}
    P_{\mathcal{B}}(s) = 
    \begin{cases} 
    1 - \lambda & \text{if } s = \mathcal{B}_{\text{draft}} \text{ (Small)} \\
    \lambda & \text{if } s = \mathcal{B}_{\text{global}} \text{ (Large)} \\
    0 & \text{otherwise}
    \end{cases}
\end{equation}
where $\mathcal{B}_{\text{draft}}$ (e.g., 4) corresponds to the drafting stage and $\mathcal{B}_{\text{global}}$ (e.g., 1024) corresponds to the revision stage.

Empirically, we find that a small fraction of global exposure is sufficient to unlock the model's revision capabilities without compromising local generation quality. We set $\lambda = 0.1$, meaning 10\% of the training samples are processed as single large blocks (effectively full-sequence masked diffusion), while the remaining 90\% use small autoregressive blocks. 
This asymmetry reflects the inference computational budget: the model performs intensive generation at the small scale, while the large scale is primarily used for sparse edits and consistency checks. 
This mixed objective also prevents the model from overfitting to the positional biases of autoregressive boundaries, making the model's performance across varying block sizes during inference more robust.

\section{Experiments}
\label{sec:experiments}

\paragraph{Setup} 
We opted to follow the baseline configuration outlined in \citet{arriola2025block} when evaluating our approach \citep{Gokaslan2019OpenWeb} on the OpenWebText (OWT) dataset.
We utilised a Transformer backbone network with 110 million parameters (12 layers, 768 hidden dimensions).
To ensure fair comparisons, we initialise the model using the official pre-trained checkpoint provided by \citet{arriola2025block}, which was pre-trained for 850,000 gradient steps at the maximum block size ($L'=L$).

From this initialization, we fine-tune the model using our \textit{Mixed-Scale Training} objective (Section \ref{sec:mixed_training}) to enable both drafting and revision capabilities. 
Distinct from the baseline recipe, which fine-tunes for 150K gradient steps on a fixed small block size, we fine-tune for only \textbf{40K gradient steps (One-quarter of 150K)}, translating to 467B tokens. This reduced training budget highlights the data efficiency of our approach.
During fine-tuning, we adopt the variance-reduction technique from \citet{arriola2025block} by adaptively learning the range of masking rates (optimizing parameters $\beta, \omega$) to minimize the gradient variance of the diffusion loss.
All other hyperparameters, including the optimizer and learning rate schedule, follow the original implementation. All of our methods utilize a draft block size of 4.

% \subsection{Likelihood evaluation [WIP]}
% \label{sec:likelihood_eval}

\subsection{Generation Quality}
\label{sec:gen_quality}

We quantitatively assess the quality of generated samples by computing perplexity (Gen PPL). As diffusion models cannot compute perplexity autonomously, this metric is derived using a pre-trained GPT-2-Large model.
Unlike the static likelihood evaluation on ground-truth data, this metric measures the coherence and fluency of the model's actual open-ended generations.
Table \ref{tab:gen_ppl} shows the unconditional generation results for multiple different models at lengths $L=1024$ and $L=2048$.

\begin{table}[htb!]
\centering
% \resizebox{\columnwidth}{!}{
\begin{tabular}{lcccc}
\toprule
& \multicolumn{2}{c}{\textbf{$L=1024$}} & \multicolumn{2}{c}{\textbf{$L=2048$}} \\
\cmidrule(lr){2-3} \cmidrule(lr){4-5}
\textbf{Model} & \textbf{Gen. PPL} & \textbf{NFEs} & \textbf{Gen. PPL} & \textbf{NFEs} \\
\midrule
AR & \textbf{\textit{14.1}} & 1K & \textbf{\textit{13.2}} & 2K \\
SEDD & 52.0 & 1K & 41.3 & 2K \\
MDLM & 46.8 & 1K & 35.3 & 2K \\
\midrule
\multicolumn{5}{l}{\textit{Prior Block Diffusion Work}} \\
SSD-LM ($L'=25$) & 37.2 & 40K & 35.3 & 80K \\
BD3-LM ($L'=16$) & 33.4 & 1K & 31.5 & 2K \\
BD3-LM ($L'=8$) & 30.4 & 1K & 28.2 & 2K \\
BD3-LM ($L'=4$) & 25.7 & 1K & 23.6 & 2K \\
 & 25.0 & 1.5K & 22.8 & 3K \\
\midrule
\midrule
\multicolumn{5}{l}{\textit{Structural Block Diffusion (Using 26\% Tuning Data)}} \\
\textbf{Ours (Stage 1 only)} & 27.4 & 1.0K & 25.1 & 2.0K \\
\textbf{Ours (Full 2-Stage)} & 24.6 & 1.1K & 22.5 & 2.2K \\
 & 22.6 & 1.2K & 21.2 & 2.5K \\
 & \textbf{21.9} & 1.5K & \textbf{20.6} & 3.0K \\
\bottomrule
\end{tabular}
% }
\caption{Generative perplexity (Gen. PPL; $\downarrow$) and number of function evaluations (NFEs) for unconditional generation of lengths $L=1024$ and $L=2048$. Generative perplexity is evaluated using GPT2-Large. Numbers for AR, SEDD, MDLM, SSD-LM and BD3-LM models are borrowed from \citet{arriola2025block}.
% While requiring more inference steps due to the two-stage process, our method achieves superior sample coherence, surpassing the best BD3-LM baseline.
}
\label{tab:gen_ppl}
\end{table}

As shown in Table \ref{tab:gen_ppl}, our "Stage 1 only" model serves as the drafting baseline with a Gen PPL of 27.4, slightly higher than BD3LM($L'=4$), given that our training budget is quartered. 
Applying the Stage 2 global revision dramatically reduces the perplexity to \textbf{21.9} (a relative improvement of $\approx 20\%$). 
This result empirically validates our core hypothesis: while small blocks are sufficient for local syntax, they suffer from "myopic" errors that accumulate over long sequences. The second stage effectively corrects these long-range inconsistencies by leveraging the global receptive field. 
Notably, this performance is achieved using only \textbf{26\% of the tuning budget} (40K steps) compared to the fully converged baselines (150K steps), highlighting the extreme data efficiency of our structural refinement approach.

% \paragraph{Comparison with Strong Baselines.}
Our method establishes a new state-of-the-art for diffusion-based models at this scale.
We outperform the strongest baseline, BD3-LM ($L'=4$), reducing Gen PPL from 25.0 (same NFEs) to \textbf{21.9} at $L=1024$.
This advantage widens for longer sequences ($L=2048$), where our method achieves a Gen PPL of \textbf{20.6} compared to the baseline's best of 22.8 (same NFEs). 
By effectively narrowing the gap to autoregressive models (14.1), Structural Block Diffusion demonstrates that the "Draft-then-Revise" paradigm significantly enhances the expressivity of diffusion architectures.

\paragraph{Quality-Efficiency Trade-off}
To address the concern of increased computation for multi-stage generations, we explicitly analyze this Quality-Speed trade-off using NFEs.
Our method offers a flexible Pareto frontier rather than a fixed cost:
\begin{itemize}
    \item \textbf{Minimal Overhead (1.1K NFEs):} With just a 10\% increase in compute over the standard baseline, our method achieves a PPL of 24.6, already surpassing the standard BD3-LM (25.7).
    \item \textbf{Iso-compute Superiority (1.5K NFEs):} When we allow the baseline BD3-LM more compute steps (1.5K) to match our re-masking budget, it only improves to 25.0. In contrast, our method reaches \textbf{21.9} under the same computational budget. This further proves that our performance gains stem from framework design rather than from increased sampling steps.
    \item \textbf{Maximum Quality (3.0K NFEs):} To get optimal performance, further extending the compute budget yields the best result of 20.6, opening up possibilities that are impossible for single-pass models.
\end{itemize}

\subsection{Ablations and Analysis}
\label{sec:ablations}

We conducted comprehensive ablation experiments to validate multiple key design choices. Unless otherwise specified, all ablation experiments were performed on the OWT dataset using the default two-stage configuration.

\subsubsection{Impact of Revision Scope (Block Size \& Ratio).}
\begin{wrapfigure}{r}{0.5\textwidth}
    \centering
    \includegraphics[width=\linewidth]{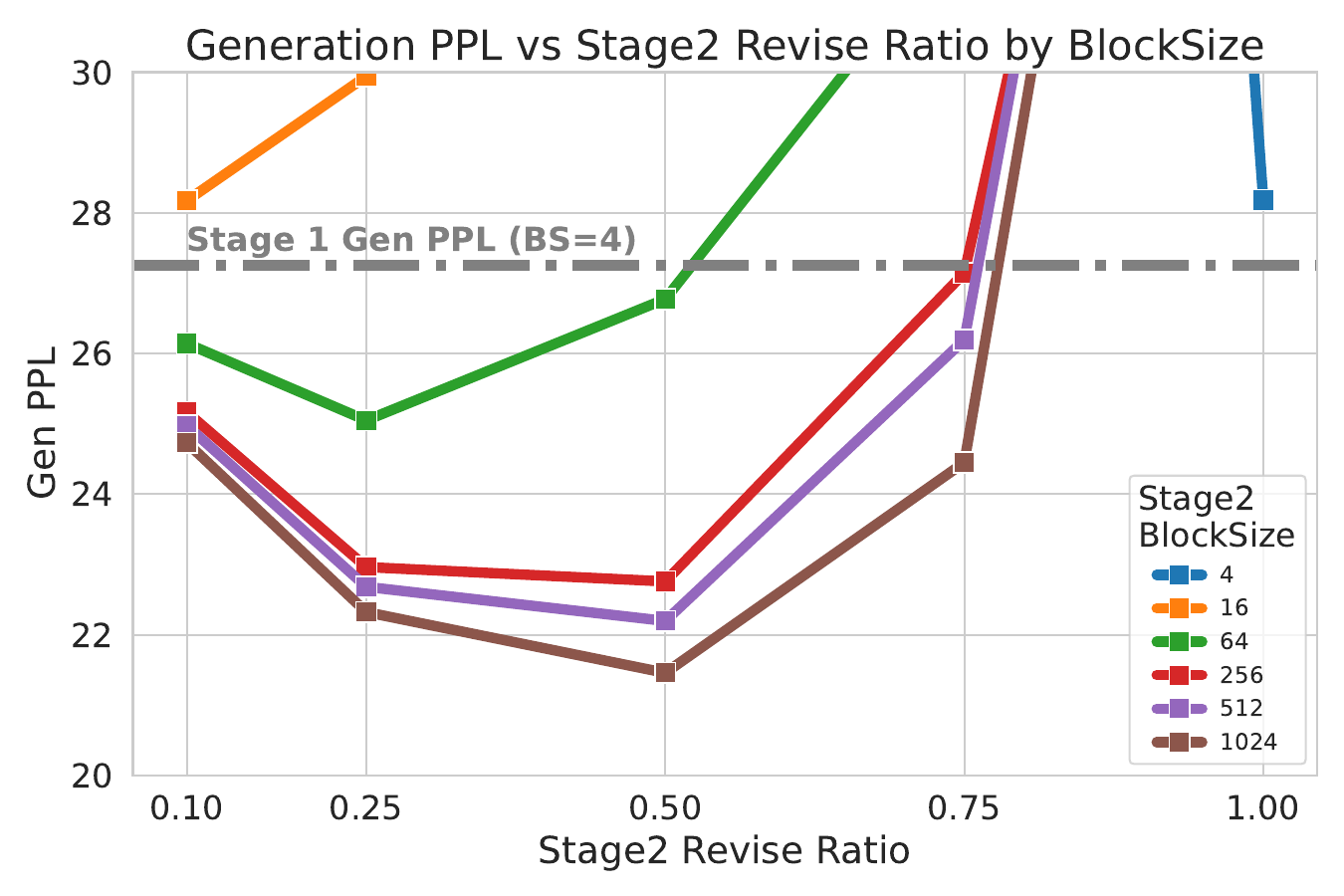}
    \caption{Ablation on Revision Scope. Generative Perplexity (Gen PPL) as a function of the Stage 2 revision ratio $\gamma$ across varying block sizes $\mathcal{B}^{(2)}$. The gray dashed line represents the Stage 1 baseline (BS=4).}    
    \label{fig:ablation_bs_ratio}
    % \vspace{-10pt} 
\end{wrapfigure}

The efficacy of the revision stage depends on two hyperparameters: the Stage 2 block size $\mathcal{B}^{(2)}$ (determining the receptive field) and the masking ratio $\gamma$ (determining the extent of modification).
Figure \ref{fig:ablation_bs_ratio} visualizes the Generative Perplexity across different Stage 2 block sizes $\mathcal{B}^{(2)}$ and revision ratios $\gamma$.

\paragraph{Global Context is Necessary}
We observe a clear performance gain as the block size increases. As $\mathcal{B}^{(2)}$ increases from 4 to 1024, the perplexity curves shift downwards monotonically. 
Employing small block sizes (e.g., $\mathcal{B}^{(2)} \in \{4, 16\}$) in the second stage fails to beat the Stage 1 baseline (the gray dashed line), and in some cases even degrades quality. 
We start to get significant gains when $\mathcal{B}^{(2)} \ge 64$, with the best performance achieved at $\mathcal{B}^{(2)}=1024$. 
This confirms that the success of the revision phase hinges entirely upon the newly added access to the global context.

\paragraph{The U-Shaped Trade-off}
Regarding the masking ratio $\gamma$, results exhibit a convex "U-shaped" trend across all effective block sizes, minimizing perplexity between 0.25 to 0.5. 
Too low ratios ($\gamma \le 0.1$) overly constrain the revision process, while too high ratios ($\gamma \ge 0.75$) degrade performance by discarding the critical \textit{structural skeleton} provided by the draft. 
Interestingly, as $\gamma \to 1.0$, performance explodes beyond the baseline, indicating the Block Diffusion model degrades to an MDLM model, demonstrating that the retention of partial drafts is crucial for stabilising the high variance inherent in purely non-autoregressive generation.

\subsubsection{Effectiveness of Remasking Strategy.}

We investigate how the token selection policy affects refinement quality using a fixed revision setting ($\mathcal{B}^{(2)}=1024, \gamma=0.5$). 
We compare our Snapshot Confidence strategy against two variants: (1) Random Masking, and (2) Post-hoc Confidence.

\begin{table}[htb!]
\centering
\begin{tabular}{lc}
\toprule
\textbf{Remasking Strategy} & \textbf{Gen. PPL ($\downarrow$)} \\
\midrule
\textit{Baseline (Stage 1 Output)} & 27.36 \\
\midrule
Random Masking & 30.26 \\
Post-hoc Confidence & 29.85 \\
\textbf{Snapshot Confidence (Ours)} & \textbf{21.85} \\
\bottomrule
\end{tabular}
\caption{Ablation on Remasking Strategy. Comparison of Generative Perplexity (Gen. PPL) using different token selection policies for the revision stage (Stage 2). All experiments use $\mathcal{B}^{(2)}=1024$ and $\gamma=0.5$. Notably, only our \textit{Snapshot Confidence} strategy improves upon the Stage 1 baseline, while other strategies degrade performance.}
\label{tab:ablation_strategy}
\end{table}

Table \ref{tab:ablation_strategy} reveals two critical insights regarding the difficulty of effective refinement:
\begin{enumerate}
    \item \textbf{Blind Masking is Hurtful} Random masking significantly degrades performance, increasing the PPL from 27.36 to 30.26. This demonstrates that indiscriminately disrupting structural frameworks yields detrimental effects: when meaningful tokens are arbitrarily removed, the model struggles to reconstruct coherent text.
    \item \textbf{Post-hoc Scores Fail} The method based on Post-hoc confidence selection also failed to surpass the baseline level (29.85). We attribute this to ‘model overconfidence’: once the sequence is fully generated, the model tends to persistently believe its prior hallucinations. This strategy likely preserves errors without providing any useful signals, while simultaneously obscuring valid but low-probability tokens, leading to further deterioration in performance.
\end{enumerate}

By contrast, our snapshot confidence strategy significantly reduces the PPL to \textbf{21.85}. By capturing the uncertainty dynamics during diffusion, this strategy successfully locks in tokens that the model initially struggles to generate, enabling the second stage to focus on correcting actual errors rather than disrupting valid structures.

\subsubsection{Training Mix Configuration.}

Finally, we justify our specific choice of the Mixed-Scale Training objective (Section \ref{sec:mixed_training}). We compare our default Bimodal distribution ($\mathcal{B} \in \{4, 1024\}$) against three alternatives:
\begin{itemize}
    \item \textbf{Baseline (No Mix):} Fine-tuning only on small blocks ($\mathcal{B}=4$).
    \item \textbf{Comparison I (Scale):} Mixing small blocks with an intermediate size ($\mathcal{B} \in \{4, 512\}$).
    \item \textbf{Comparison II (Complexity):} A uniform mixture of all power-of-two scales $\{4, 16, 64, \dots, 1024\}$.
\end{itemize}

\begin{table}[htb!]
\centering
\begin{tabular}{lcc}
\toprule
\textbf{Training Configuration} & \textbf{Stage 1 PPL} & \textbf{Stage 2 PPL} \\
\midrule
\textit{Baseline (No Mix)} & 27.95 & 31.97 \textcolor{red}{$\uparrow$} \\
\midrule
Uniform Mixture ($\{4, 16, \dots, 1024\}$) & 31.98 & 22.60 \\
\midrule
Bimodal Mix (4, 512) & \textbf{27.26} & \textbf{21.46} \\
Bimodal Mix (4, 1024) & 27.36 & 21.85 \\
\bottomrule
\end{tabular}
\caption{Ablation on Training Mix Strategy. Evaluating the impact of the Mixed-Scale Training objective. We report the PPL for the initial draft (Stage 1, $\mathcal{B}=4$) and the final revision (Stage 2, $\mathcal{B}=1024$). The "Baseline" is a naively Block-Diffusion trained model.}
\label{tab:ablation_training}
\end{table}

Table \ref{tab:ablation_training} summarizes the perplexity at both drafting (Stage 1) and revision (Stage 2) phases. It first shows that mixed training is necessary for the framework. The Baseline model, trained only with the vanilla Block Diffusion Training, completely fails at the revision stage (PPL degrades from 27.95 to 31.97). This confirms that the standard block diffusion model cannot be generalised to global scenarios under zero-shot conditions, and explicit training of large blocks is a prerequisite for achieving structural refinement.
    
\paragraph{Bimodal vs. Uniform} The complex "Uniform Mixture" strategy performs poorly in the drafting phase (Stage 1 PPL 31.98 vs. 27.36 for Bimodal). We attribute this to an overly intrusive optimisation process: excessively granular training block levels undermine the model's foundational capabilities, resulting in low-quality drafts that prove more challenging to refine. Bimodal training delivers clearer, more explicit multi-task signals.
    
\paragraph{Optimal Revision Scale} Comparing the bimodal strategies, $\text{Mix}(4, 512)$ and $\text{Mix}(4, 1024)$ perform comparably, with the 512-mix achieving marginally lower perplexity (21.46 vs. 21.85). While intermediate scales are easier to optimize, we select $\text{Mix}(4, 1024)$ as our default to ensure the model is theoretically capable of strictly full-sequence modeling without windowing artifacts, securing robust performance for the target evaluation length of $L=1024$.

\section{Related Works}
\label{sec:related_works}

\paragraph{Discrete Diffusion Language Models}
Discrete diffusion models have now emerged as strong competitors to autoregressive paradigms. They are renowned for their non-autoregressive parallel generation mechanism, which offers greater controllability and flexibility than traditional AR models.
Early explorations like D3PM \citep{austin2021structured} modeled the forward noise process using a transition matrix that extended the Gaussian diffusion model to a discrete state space.
Subsequently, masked discrete diffusion models \citep{Gu_2022_CVPR, li2022diffusion} gained prominence by explicitly incorporating a mask token to assist in reversing data corruption.
Recently, approaches like SEDD \citep{lou2024discrete} and MDLM \citep{sahoo2024simple} unified discrete diffusion processes with continuous time steps. By modulating noise timestep into the model, they further reduced the language modeling perplexity of diffusion models.
Recent attempts to scale masked discrete diffusion models to 7B-scale \citep{nie2025large, ye2025dream} and beyond have demonstrated advantages in planning, mathematical, and coding tasks, further expanding their potential.
The core architectural advantage of these models lies in their \textit{global bidirectional attention mechanism}, which enables exceptional performance in global text structure planning and infilling tasks. However, this mechanism precludes the use of key-value caching techniques, resulting in quadratic inference complexity with sequence length. This limitation severely restricts their applicability in long-context generation tasks.

\paragraph{Semi-Autoregressive Block Generation}
To address the efficiency challenges, researchers proposed the semi-autoregressive paradigm. Characterized by autoregressive behavior at the macro level and diffusion at the micro level, this approach leverages KV Cache to reduce computational while preserving a certain degree of local bidirectional parallelism.
SSD-LM \citep{han2023ssd} initially introduced the semi-AR diffusion model, achieving linear inference complexity through iterative text block generation.
Subsequently, the Block Discrete Denoising Diffusion Model \citep{arriola2025block} further optimized this paradigm. It employs inter-block autoregression and intra-block bidirectional diffusion, enabling KV cache utilization while allowing flexible control over generation length.
However, this macro-level autoregressive approach also introduces inherent compromises. In semi-AR models, blocks can be abstracted as metatokens. These metatokens are constrained by strict unidirectional and irreversible dependencies, resulting in the loss of diffusion models' inherent global planning capabilities. Moreover, early stage generation errors become irrevocable and cannot be corrected based on subsequent context, inevitably compromising consistency in long-text generation.

\paragraph{Iterative Refinement and Multi-Stage Generation}
Our approach is also closely related to the "Draft-then-Revise" paradigm found in multi-stage generation. 
In the AR domain, Speculative Decoding \citep{leviathan2023fast, chen2023accelerating} leverages a small model to rapidly draft tokens and a large model to verify them, aiming to accelerate inference without degrading quality. 
More fundamentally, there are models like the Levenshtein Transformer \citep{gu2019levenshtein} optimize text through iterative insertion and deletion operations. 
While Speculative Decoding mainly focuses on speeding up generation, our proposed Diffusion in Diffusion focuses on improving quality by reintroducing global planning and coherence to BD3-LMs. 
Rather than simply rejecting tokens based on likelihood, we use snapshot confidence to identify structural defects and bring back the global receptive field for repair. 
Fundamentally, our approach is a nested iterative diffusion process carefully designed to restore the long-range modeling capabilities of global diffusion within an efficient block-based diffusion framework.

\section{Conclusion}

This work aims to address the fundamental trade-off of global coherence and  inference efficiency in semi-autoregressive diffusion models. 
We propose \textsc{Diffusion in Diffusion}, a structural diffusion framework that improves both the global coherence and the generation quality of existing block diffusion models with minimal overhead.
This approach introduces a multi-stage refinement mechanism based on snapshot confidence and a mix-scale training strategy, re-injecting the long-range planning capability of Global Diffusion into an efficient block-generation framework. 
Results demonstrate that this Draft-then-Revise paradigm achieves state-of-the-art generation perplexity performance on OpenWebText with exceptional data efficiency, significantly narrowing the performance gap with traditional autoregressive models.

\newpage

\bibliography{colm2026_conference}
\bibliographystyle{colm2026_conference}

% \appendix
% \section{Appendix}

\end{document}